%% file: root.tex
\documentclass[letterpaper, 10 pt, conference]{ieeeconf}  

\IEEEoverridecommandlockouts                              

\let\labelindent\relax

\usepackage{graphics} 
\usepackage{epsfig} 
\usepackage{times} 
\usepackage{amsmath} 
\usepackage{amssymb}  

\usepackage{microtype}
\usepackage{graphicx}
\usepackage{subfig} 
\usepackage{booktabs} 
\usepackage{bm}
\usepackage{url}
\usepackage{enumitem}
\setlist[enumerate]{topsep=0pt,itemsep=-0.5ex,partopsep=1ex,parsep=1ex}
\usepackage{todonotes}

\usepackage{hyperref}

\usepackage{algorithm}
\usepackage{algorithmic}
\usepackage{array}
\usepackage{multirow}

\newcommand\MyBox[2]{
  \fbox{\lower0.75cm
    \vbox to 1.7cm{\vfil
      \hbox to 1.7cm{\hfil\parbox{1.4cm}{#1\\#2}\hfil}
      \vfil}%
  }%
}
\title{\LARGE \bf
Renaissance Robot: Optimal Transport Policy Fusion \\for Learning Diverse Skills
}

\author{Julia Tan$^{1}$, Ransalu Senanayake$^{2}$ and Fabio Ramos$^{3}$
\thanks{*This work was not supported by any organization}
\thanks{$^{1}$Julia Tan is with the School of Computer Science,
        University of Sydney, Sydney, Australia
        {\tt\small jtan5838@uni.sydney.edu.au}}%
\thanks{$^{2}$Ransalu Senanayake is with the Dept. of Computer Science, Stanford University,
        Stanford, CA 94305, USA
        {\tt\small ransalu@cs.stanford.edu}}%
\thanks{$^{3}$Fabio Ramos is with NVIDIA, USA and School of Computer Science,
        University of Sydney, Australia
        {\tt\small ftozetoramos@nvidia.com}}%
}

\begin{document}

\maketitle
\thispagestyle{empty}
\pagestyle{empty}

\begin{abstract}
Deep reinforcement learning (RL) is a promising approach to solving complex robotics problems. However, the process of learning through trial-and-error interactions is often highly time-consuming, despite recent advancements in RL algorithms. Additionally, the success of RL is critically dependent on how well the reward-shaping function suits the task, which is also time-consuming to design. As agents trained on a variety of robotics problems continue to proliferate, the ability to reuse their valuable learning for new domains becomes increasingly significant. In this paper, we propose a post-hoc technique for policy fusion using \emph{Optimal Transport} theory as a robust means of consolidating the knowledge of multiple agents that have been trained on distinct scenarios. We further demonstrate that this provides an improved weights initialisation of the neural network policy for learning new tasks, requiring less time and computational resources than either retraining the parent policies or training a new policy from scratch. Ultimately, our results on diverse agents commonly used in deep RL show that specialised knowledge can be unified into a “Renaissance agent", allowing for quicker learning of new skills.

\end{abstract}


\section{INTRODUCTION}
\input{intro}

\section{RELATED WORK}
\input{background_v2}

\section{OPTIMAL TRANSPORT POLICY FUSION}
\input{ot_fusion}

\section{EXPERIMENTS}
\input{experiments}


\section{CONCLUSION}
\input{conclusion}

\addtolength{\textheight}{-12cm}   





\section*{ACKNOWLEDGMENT}

We thank Anthony Tompkins for early discussions and ideas, especially on Optimal Transport theory.



\bibliography{root}
\bibliographystyle{ieeetr}

\end{document}

%% file: intro.tex
Intelligent systems are increasingly penetrating into real-world applications, incurring the need for sophisticated models that can perform complex and adaptive tasks. Deep neural networks are already ubiquitous in solving classification problems in many fields of interest. Recent advancements in reinforcement learning (RL) have extended the capacity of neural networks into the physical domain, allowing for intelligent control of robotic agents in a dynamic environment. Accordingly, RL in robotics has the potential to advance key industries, from manufacturing \cite{manufacturing} to healthcare \cite{healthcare}, agriculture \cite{agriculture} or transportation \cite{transport}.

While RL is progressively being used in a range of robotics problems, most real-world applications require numerous specialised capabilities. The computational resource constraints associated with sample-inefficient training, combined with the complexity of re-designing the appropriate reward structures for each capability, make it difficult to achieve solutions for varying goals in a reasonable timeframe. Traditional methods of transfer learning are also inadequate, without a robust way to determine which parts of a policy are still relevant in the new context. Even with targeted policy reuse, the knowledge of only one agent can be leveraged. This is likely not sufficient for most cases, since small differences in the environment or the kinematics of the robotic agent itself can drastically change the rewards of previously familiar actions. Consequently, we require a minimal dependency fusion method that can effectively consolidate the learned behaviours of multiple parent policies. 

\begin{figure}[t!]
    \centering
    \includegraphics[width=\linewidth]{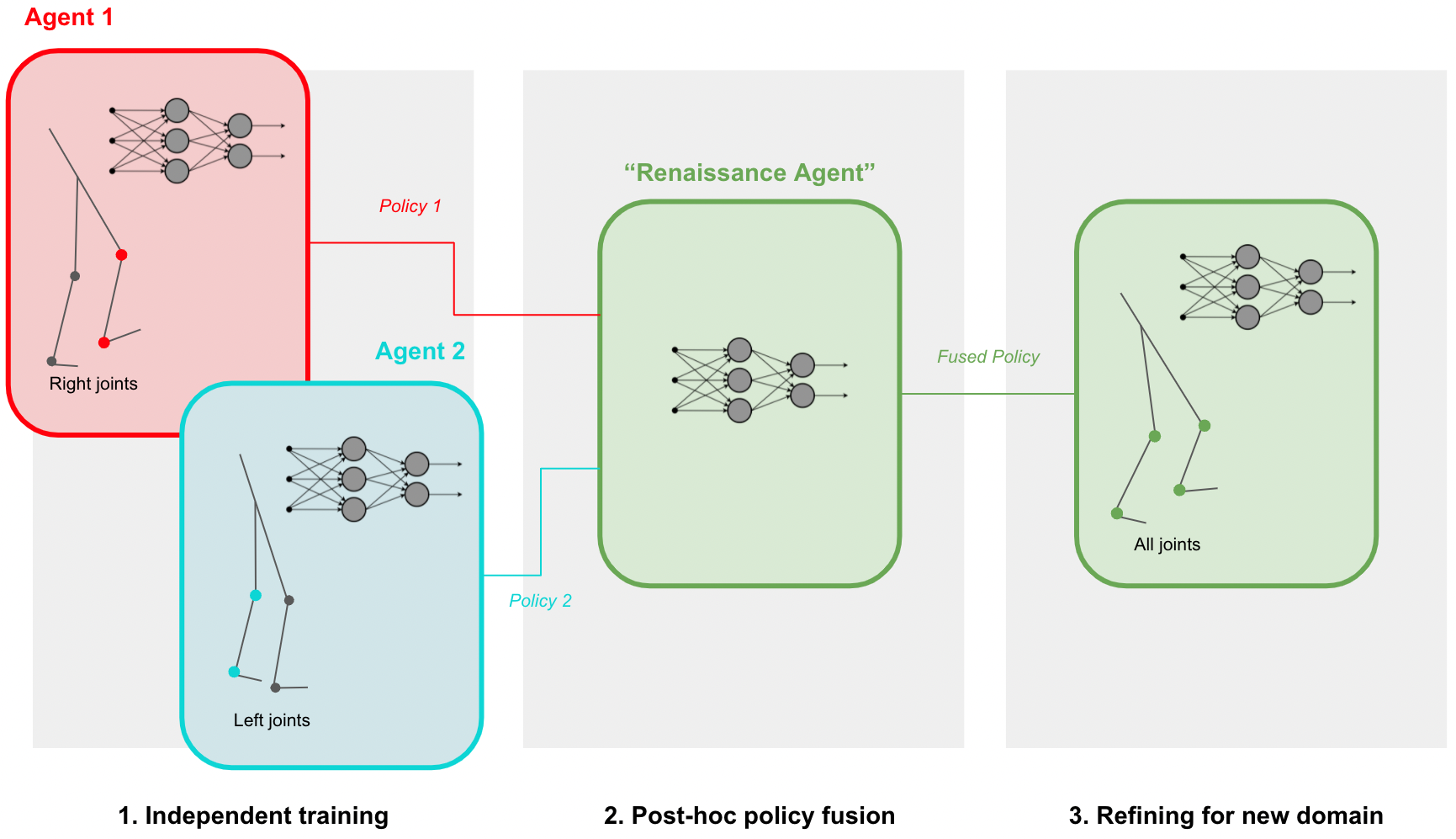}
    \caption{First, two policy networks are trained independently on an RL robotics task, where either the agent or environment has been altered in distinct ways (e.g. for the Gym environment Walker2d, using primarily right joints vs left joints). Next, these policy networks are fused post-hoc to produce an intialisation of weights for the Renaissance agent. Finally, the Renaissance agent is refined on a new domain (e.g. walking using all joints) to learn new skills beyond the original capabilities of the parent agents.}
    \label{fig:summary}
\end{figure}

In this paper, we present a technique to fuse RL policies represented as neural networks using Optimal Transport (OT), and propose the fused policy as a superior initialisation of weights for faster learning of diverse skills. With reference to the Renaissance man, we call this fusion result a ``Renaissance agent".

In particular, our contributions are:
\begin{enumerate}
    \item A method based on OT for efficiently combining RL policies post-hoc, requiring no awareness of the original data used to train them;
    \item A demonstration that this OT fusion approach generalises the prior knowledge of parent policies over different environments as well as different robot kinematics, and
    \item Validation that the Renaissance agent learns new skills more effectively and within a shorter time period than either retraining the parent agents or training a new agent from scratch.
\end{enumerate}

%% file: background_v2.tex
Information fusion is a  widely studied topic in robotics for decades~\cite{durrant1990sensor}. As opposed to the common paradigm of controlling multiple robots using a centralised controller~\cite{sharma2021survey} or fusing data from various sources~\cite{alatise2020review}, our objective is to gather data from individually operating robots each with a different skill to train a single robot that possesses diverse skills appropriately learned from all other robots by fusing knowledge. Roboticists have attempted to perform knowledge fusion at the perception stage or decision-making stage of the robot autonomy stack. 
Most research studies in the perception level focuses on either fusing data from multiple sensors or at the perception algorithm level. The former, sensor fusion, is a well-studied topic in robotics~\cite{durrant1990sensor,alatise2020review} and has applications ranging from field robotics~\cite{durrant1990sensor,alatise2020review} to wearable sensors~\cite{novak2015survey}. There are also studies on aggregating data from multiple robots~\cite{guo2018multirobot} as the simplest fusion technique. Some of the main limitations of such aggregation methods include the requirement of saving the ever growing dataset and training the model online. 

While there are many techniques for data fusion, fewer exist for fusing learned perception models. This is because it is more difficult to determine how and what to fuse for neural networks. Zhi et al.~\cite{zhi2019continuous} has proposed a technique to fuse Bayesian Hilbert maps from data gathered from various agents. However, it is not clear how the proposed occupancy map fusion technique can be used when the underlying model is a neural network. 

Researchers have studied model fusion techniques for various applications beyond robotics. The most common technique, ensembling~\cite{ensembling1,ensembling3} requires all models to be running concurrently, which is computationally expensive. Moreover, it does not actually aggregate each model's distinct capabilities, since decision making occurs separately. Hence, it is not suitable for adapting to customised tasks, especially when the learned behaviours of the parent models differ significantly. In contrast, there are models that result in a single model by transferring knowledge from a teacher network~\cite{distillation1}. Whilst there are also techniques~\cite{network_fusion1} that yield a combined model without the need for retraining, they also do not perform empirically better than naive averaging. More recently, in federated learning settings, aligning neural networks has been studied~\cite{alignment1}. In such frameworks, the number of parameters required grows with the size of the data, so it is infeasible for fusing large data models ubiquitous in robotics. More recently, others have used OT theory to fuse neural network models~\cite{model_fusion_via_ot}, although this has not yet been applied to RL or decision-making. When fusing robots, it is ideal if we can make them as agnostic as possible to sensory settings because the ultimate goal of a robot is to maneuver safely and efficiently.

On the other end of the fusion spectrum, it is possible to combine decision-making modules~\cite{Dax2022icra} or their outputs, though the latter, by construction, is ill-informed about data and typically produces less accurate results. Researchers have investigated how to combine the outputs of a traditional hand-crafted controller with a deep RL controller using Bayesian inference~\cite{rana2021bayesian}. However, it is not clear how to combine two or more neural network controllers, especially at the decision-making stage rather than once the outputs are produced. In our work, we fuse architecturally similar deep RL controllers that possess different skills. Though Dax et al.~\cite{Dax2022icra} also discusses sensor selection and decision fusion, their proposed model also does not represent policies using neural networks.

%% file: ot_fusion.tex
\subsection{Policy Networks in RL}

In RL, an agent is informed by a policy $\pi: \mathcal{A} \times \mathcal{S} \to [0,1]$ that determines what actions $\mathcal{A}$ to perform given the current states $\mathcal{S}$. A policy $\pi_\theta^\mathcal{E}$, trained in domain $\mathcal{E}$, can be represented as a neural network parameterized by $\theta$. When training a policy, an RL algorithm such as Proximal Policy Optimisation (PPO) \cite{ppo} or Deep Deterministic Policy Gradient (DDPG) \cite{ddpg} is used to update the weights of the neural network so that the cumulative reward is maximised over time. Depending on the task configuration during training, these learned weights will differ between agents. However, for policy networks that have the same input features, similar weight vectors at each individual layer may represent common learned patterns. Therefore, we wish to find an efficient way to consolidate the common patterns while diminishing any over-optimised patterns, so that we can generate a Renaissance agent that is capable of leveraging prior knowledge for new domains.

For two policy networks, $\pi_{\theta_1}^{\mathcal{E}_1}, \pi_{\theta_2}^{\mathcal{E}_2}$, trained independently in domains, $\mathcal{E}_1, \mathcal{E}_2$, respectively, our objective is to determine the best fusion operation $c: \Pi^{\mathcal{E}_1} \times \Pi^{\mathcal{E}_2} \to \Pi^{\mathcal{E}_*}$ for the policies $\pi^\mathcal{E} \in \Pi^\mathcal{E}$ and new domain ${\mathcal{E}_*}$ such that empirically $r(\pi_{\theta_*}^{\mathcal{E}_*}) > r(\pi_{\theta_1}^{\mathcal{E}_*}), r(\pi_{\theta_2}^{\mathcal{E}_*})$, where $r(\pi^\mathcal{E})$ is the reward when policy $\pi$ is deployed in domain $\mathcal{E}$. For instance, naive policy averaging $c(\pi_{\theta_1}^{\mathcal{E}_1},\pi_{\theta_2}^{\mathcal{E}_2}) = (\pi_{\theta_1}^{\mathcal{E}_1}+\pi_{\theta_2}^{\mathcal{E}_2})/2$ is a simple fusion operation. However, it is ineffective since it does not find the optimal alignment between the parameter vectors of the policies. In this paper, we propose that OT theory is apt for finding an alignment map between the weights of two policy networks. This enables us to match the weights that are most likely to represent common patterns.

\subsection{Optimal Transport Theory}

\subsubsection{Monge-Kantorovich Problem}
OT theory has previously been applied to mapping problems in order to allow robots to operate in new environments \cite{Tompkins2020rss}. However, in this work, we utilise OT for policy fusion so that robotic agents can consolidate learned behaviours for new domains, including new environments as well as new kinematics. Typically, OT is used to transform a single source distribution $\mathcal{S}$ to a target distribution $\mathcal{T}$ \cite{ot}. When applied to the incoming weight vectors of a policy network, OT represents them as distributions of Dirac ``masses" \cite{ot_masses}. The ``distance" between any two distributions is then quantified by the least costly way of transporting each mass from $\mathcal{S}$ to $\mathcal{T}$. 

According to the Monge-Kantorovich Problem \cite{monge_kant}, for the datasets $\bm{\mu}^{(\mathcal{S})}$ and $\bm{\mu}^{(\mathcal{T})}$ of size $N^{\mathcal{S}}$ and size $N^{\mathcal{T}}$ on independent metric spaces $\Omega^{(\mathcal{S})}$ and $\Omega^{(\mathcal{T})}$, there exists an optimal probabilistic coupling $P_* \in \mathbb{R}^{N^{(\mathcal{S})}\times N^{(\mathcal{T})}}$ between the two datasets. This optimal coupling always exists for a distance function $D: \Omega^{(\mathcal{S})}\times \Omega^{(\mathcal{T})}\rightarrow [0,\infty)$, and is defined by,
\begin{equation}\label{eqn:ot_map}
  P_* = \small{\mathop{\text{arginf}}_{P\in\Gamma}}\int_{\Omega^{(\mathcal{S})}\times \Omega^{(\mathcal{T})}} D(\bm{\mu}^{(\mathcal{S})},\bm{\mu}^{(\mathcal{T})})\text{d}P(\bm{\mu}^{(\mathcal{S})},\bm{\mu}^{(\mathcal{T})}).
\end{equation}

\subsubsection{Earth Mover's Distance}
The minimum cost of transporting each mass from $\mathcal{S}$ to $\mathcal{T}$ is based on the Wasserstein metric, also known as the Earth Mover's Distance \cite{emd},
\begin{equation}
    \text{EMD}(\mathcal{S}, \mathcal{T}) = \frac{\Sigma_{i=1}^{m}\Sigma_{j=1}^{n} P_{*_{i,j}}d_{i,j}}{\Sigma_{i=1}^{m}\Sigma_{j=1}^{n} P_*{_{i,j}}},
\end{equation}
where $P_*$ is the optimal coupling and $d_{i,j}$ is the ground distance between distributions $\mathcal{S}_i$ and $\mathcal{T}_j$. We can formulate OT as an alignment problem by considering ``similarity" as inversely proportional to the Earth Mover's Distance. By computing the optimal coupling $P_*$ between two sets of weight vectors (each representing neurons from a different policy network), we are able to match the second set as the source distribution $\mathcal{S}$ to the first set as the target distribution $\mathcal{T}$ by taking the dot product,
\begin{equation}\label{eqn:ot_alignment}
    \mathcal{S}_{\text{aligned}} = P_* \cdot \mathcal{T}. 
\end{equation}

\subsection{Neural Matching and Weights Fusion}\label{sec:ot_fusion}
In order to fuse two parent policies parameterised by neural networks, we first align their neurons layer-wise so that the associated weight vectors are matched. Given that the input layers of each model receive the exact same format of input vectors, we begin neural matching from the incoming weights to the first hidden layer. As we do not know explicitly what each neuron has been trained to identify, we can pair them together based only on the ``similarity" of the incoming weights. 

A one-to-one similarity is computed by minimising the Earth Mover's Distance between the weight vectors in one model and the weight vectors in the other model as probability distributions. Hence, the optimal coupling $P_*$ represents the alignment matrix for re-positioning the weight vectors in the second model (i.e. the source distribution $\mathcal{S}$) according to the weight vectors in the first model (i.e. the target distribution $\mathcal{T}$). Taking the dot product of $P_*$ and $\mathcal{T}$ yields the aligned $\mathcal{S}$. 

\begin{figure}[H]
    \centering
    \includegraphics[width=\linewidth]{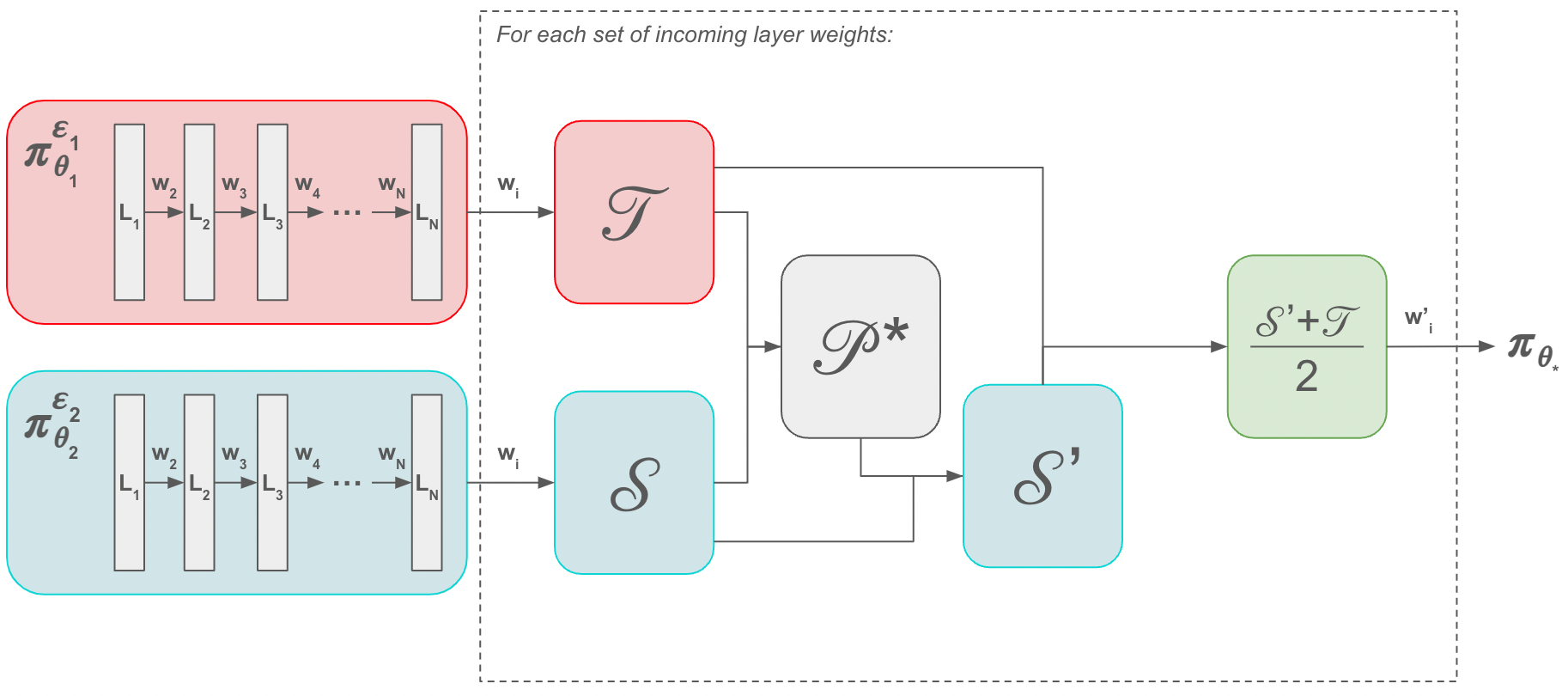}
    \caption{Our approach considers the incoming weight vectors of each independently trained policy $\pi_{\theta_1}^{\mathcal{E}_1}$ and $\pi_{\theta_2}^{\mathcal{E}_2}$. The weight vectors are considered as source $\mathcal{S}$ and target $\mathcal{T}$ distributions for OT. The result at each layer is taken as the weight vector for the fused policy.}
    \label{fig:ot_technique}
\end{figure}

To perform neural matching in the subsequent layer, we must start with rearranging the elements of each weight vector to correspond with the new positions of the neurons in the previous layer (after alignment). For the current layer, we then compute the OT map as before and align the weight vectors in the second model to the weight vectors in the first model. Algorithm \autoref{alg:align_neurons} outlines this process. Once every layer has been aligned, we then fuse the matched weights across the two policies by averaging the values. This keeps the weights with similar matches close to the same value, while changing those that are dissimilar and therefore unlikely to represent common patterns. Thus, we build the fused policy layer-by-layer, as shown in \autoref{fig:ot_technique}.
\begin{algorithm}[H]
  \caption{\emph{align\_neurons}}
  \label{alg:align_neurons}
\begin{algorithmic}
  \STATE {\bfseries Input:} weights of $N$-layer parent policies 
  \FOR{$i=2$ {\bfseries to} $N$}
    \STATE{$\mathcal{S} \leftarrow$ incoming weights to $i$-th layer of second policy}
    \STATE{$\mathcal{T} \leftarrow$ incoming weights to $i$-th layer of first policy}
    \IF{$i > 2$}
        \STATE{Rearrange $\mathcal{S}$ to match positions of previous layer alignment}
    \ENDIF
    \STATE{Compute OT map $P_*$, according to \autoref{eqn:ot_map}}
    \STATE{Align source weights to target weights using $P_*$, according to \autoref{eqn:ot_alignment}}
  \ENDFOR
  \STATE {\bfseries Output:} aligned weights of $N$-layer parent policies
\end{algorithmic}
\end{algorithm}

%% file: experiments.tex
\begin{table*}[t!]
\centering
\caption{Neural network architecture for each model.}
\label{tab:nn_architecture}
\resizebox{\textwidth}{!}{%
\begin{tabular}{@{}ccccccccccc@{}}
\toprule
 & \multicolumn{2}{c}{MNIST Binary Classifier} & \multicolumn{2}{c}{Reacher Task} & \multicolumn{2}{c}{Walker2d Task} & \multicolumn{2}{c}{HalfCheetah Task} & \multicolumn{2}{c}{FetchReach Task} \\ \midrule
Layer & Nodes & Weights & Nodes & Weights & Nodes & Weights & Nodes & Weights & Nodes & Weights \\
Input & 784 & - & 9 & - & 22 & - & 26 & - & 16 & - \\
Hidden 1 & 32 & 32 $\times$ 784 & 64 & 64 $\times$ 9 & 64 & 64 $\times$ 22 & 64 & 64 $\times$ 26 & 64 & 64 $\times$ 16 \\
Hidden 2 & - & - & 64 & 64 $\times$ 64 & 64 & 64 $\times$ 64 & 64 & 64 $\times$ 64 & 64 & 64 $\times$ 64 \\
Output & 2 & 2 $\times$ 32 & 2 & 2 $\times$ 64 & 6 & 6 $\times$ 64 & 6 & 6 $\times$ 64 & 4 & 4 $\times$ 64 \\ \bottomrule
\end{tabular}%
}
\end{table*}

\subsection{Proof of Concept: Image Classification Task}
\subsubsection{Experimental Setup}
To validate the theoretical concept behind the proposed OT fusion method, we first test it on a image classification task before extending to robotics tasks in RL. The task is constructed using the MNIST handwritten digits dataset \cite{mnist}. We independently train two parent models to be binary classifiers on the MNIST dataset. The first model classifies 0s versus all other digits, while the second model classifies 1s versus all other digits. Hence, we modify two distinct training datasets, where the class labels in each correspond to either ``yes" for the digit to be identified or ``no" for all other digits. 

There are originally 60,000 images in the training dataset. However, since the original MNIST classification problem is to identify each digit from 0 to 9, we need to rebalance the dataset in order to achieve good training outcomes for the binary classifiers. Otherwise, the models will be biased towards the class ``no", since the majority of training examples are not the digit to be identified. We use a random undersampler to rebalance the dataset so that the majority class (i.e. ``no") will have twice as many samples as the minority class (i.e. ``yes"). 

Each parent model will have ``fully-connected" layers according to the neural network architecture shown in \autoref{tab:nn_architecture}. Both models will be trained for 10 epochs on batch sizes of 128 with a 10\% validation split. Evaluating on the testing dataset of 10,000 examples (with the labels similarly modified), we get the test accuracy and test loss of each model shown in \autoref{tab:classification_parent_accuracy}.
\begin{table}[H]
\centering
\caption{Evaluation results for trained parent models.}
\label{tab:classification_parent_accuracy}
\begin{tabular}{@{}ccc@{}}
\toprule
Model & Test accuracy (\%) & \multicolumn{1}{l}{Test loss (\%)} \\ \midrule
0s                & 98.1              & 10.7                              \\
1s                & 99.0              & 8.32                             \\ \bottomrule
\end{tabular}%
\end{table}

We now reformulate the problem domain so that neither one of the parent models are able to perform well without retraining. The new task that we aim to solve is classifying both 0s and 1s versus all other digits. We fuse the parent models by applying the OT technique explained in \autoref{sec:ot_fusion}.

\subsubsection{Results}

\begin{figure}[H]
    \centering
    \includegraphics[width=0.85\linewidth]{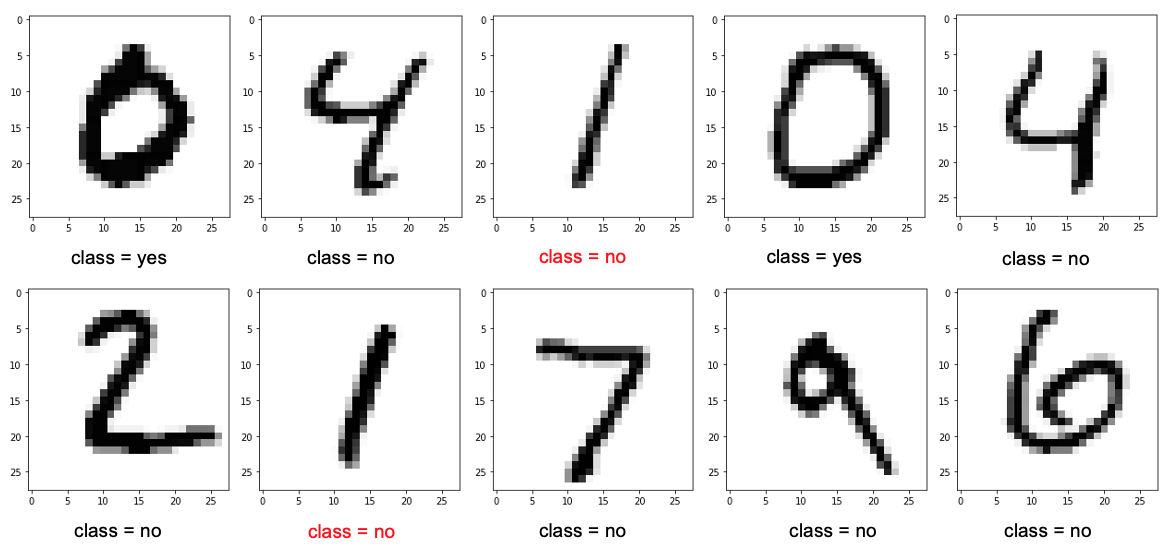}
    \caption{Example predictions of fused model without retraining, showing false negatives in red.}
    \label{fig:class_predictions}
\end{figure}

Without additional retraining, the fused model is immediately able to classify some digits correctly but not all. An example subset of the fused model's predictions is shown in \autoref{fig:class_predictions}.

The fused model performs significantly better after retraining, correctly classifying most examples. This is evident from the confusion matrix in \autoref{fig:confusion_matrix}, showing that the fused model mainly predicts true positives and true negatives.

\begin{figure}[H]
    \centering

    \begin{tabular}{l|l|c|c|c}
    \multicolumn{2}{c}{}&\multicolumn{2}{c}{Predicted label}&\\
    \cline{3-4}
    \multicolumn{2}{c|}{}&Yes&No&\multicolumn{1}{c}{}\\
    \cline{2-4}
    \multirow{2}{*}{True label}& Yes & $0.99$ & $0.01$ \\
    \cline{2-4}
    & No & $0.07$ & $0.93$ \\
    \cline{2-4}
\end{tabular}

    \caption{Normalised confusion matrix for fused model after retraining, as evaluated on testing dataset of 10,000 examples.}

    \label{fig:confusion_matrix}
\end{figure}

\begin{figure}[H]
    \centering
    \subfloat[Training accuracy.]{
    \includegraphics[width=0.85\linewidth]{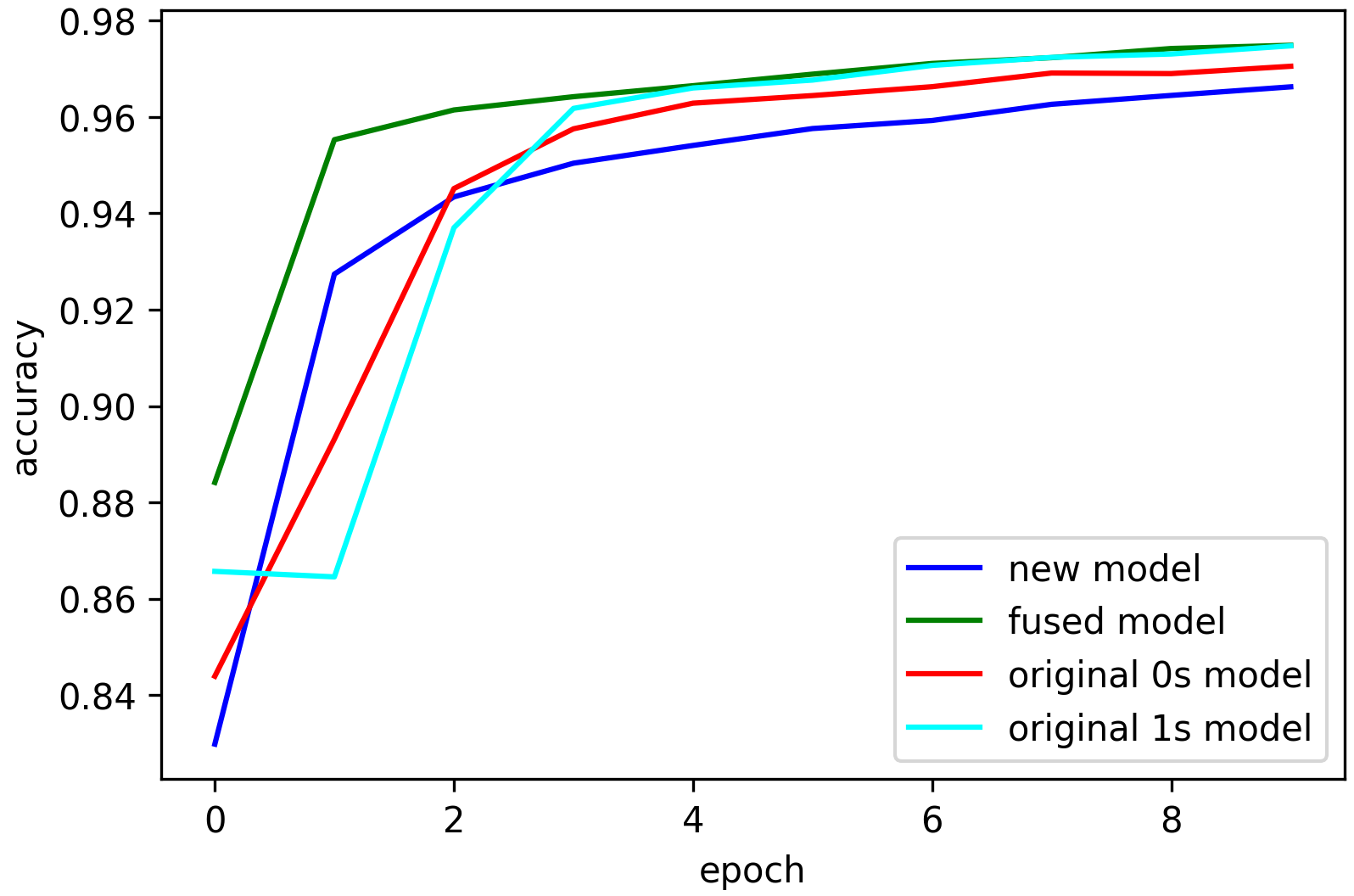}}
    \quad
    \subfloat[Validation accuracy.]{\includegraphics[width=0.85\linewidth]{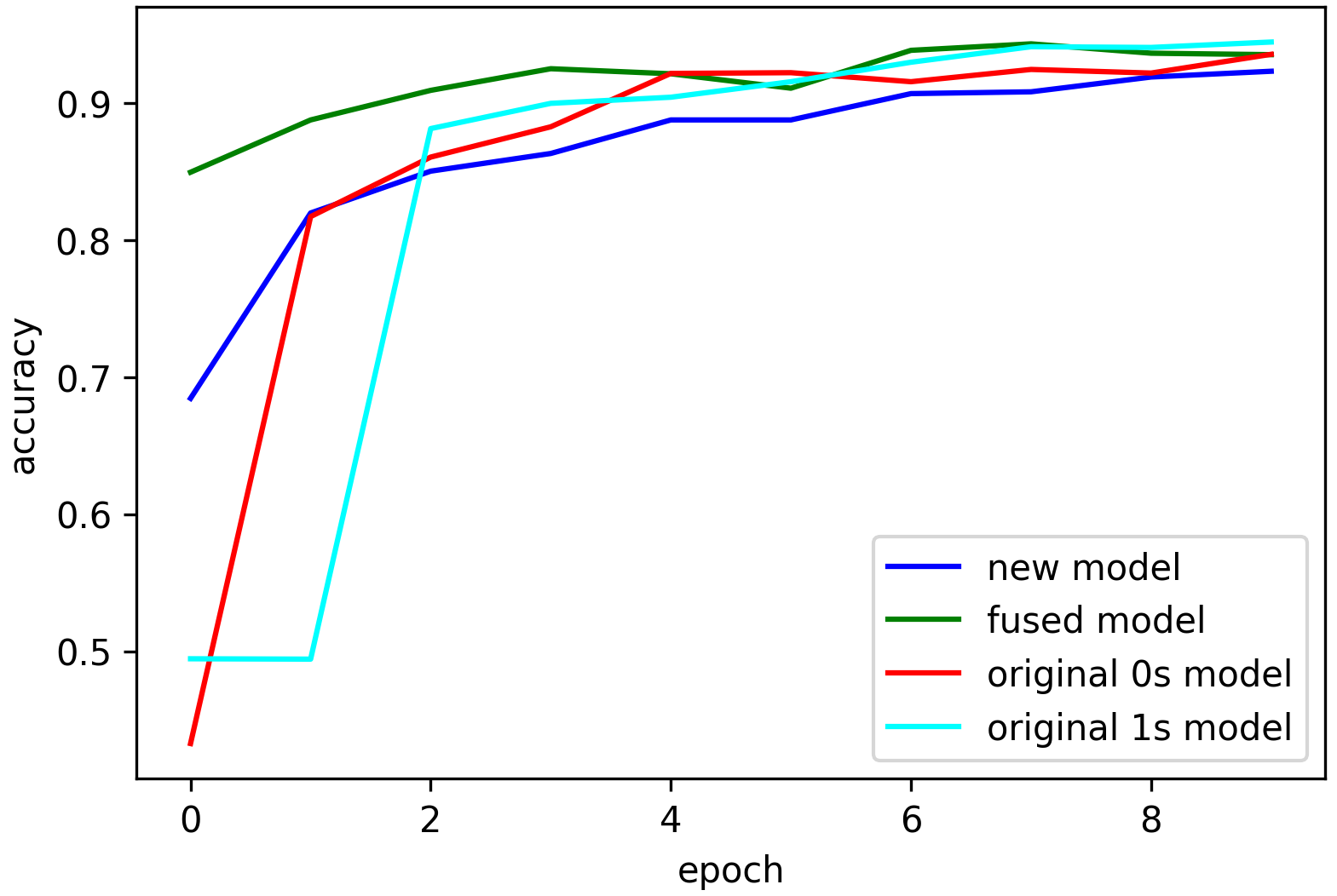}}
    \caption{Accuracy metrics over 10 retraining epochs.}
    \label{fig:class_accuracy}
\end{figure}

To evaluate the effectiveness of the fused model, we compare it with retraining the parent models as well as training a new model from scratch. The training accuracy and validation accuracy for each of these models over 10 training epochs are shown in \autoref{fig:class_accuracy}. It is evident that the fused model reaches high consistent performance much faster than the other models, surpassing a training accuracy of 95\% and a validation accuracy of 90\% after only 2 epochs. In contrast, it takes at least 4 epochs for the next best model to converge to similar performance.

\begin{figure*}[t!]
    \centering
    \subfloat[Reacher]{
    \includegraphics[width=0.2\linewidth]{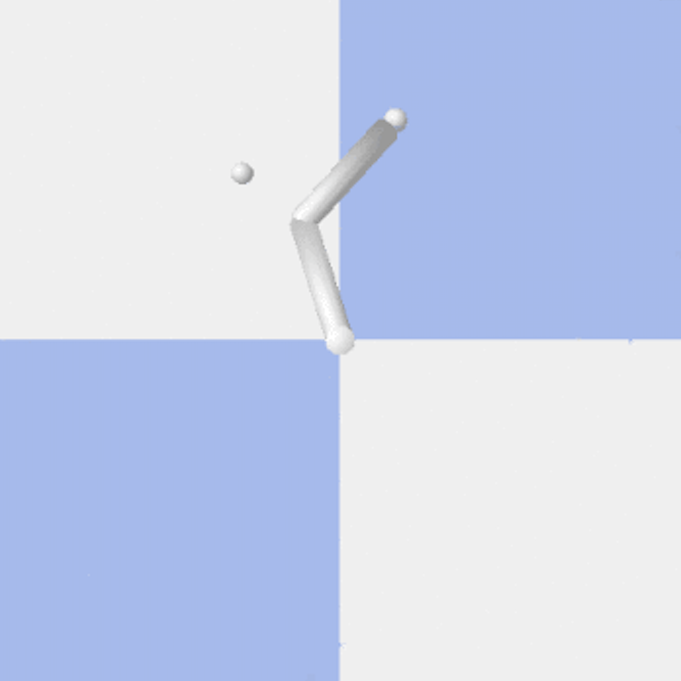}}
    \qquad
    \subfloat[Walker2d]{\includegraphics[width=0.2\linewidth]{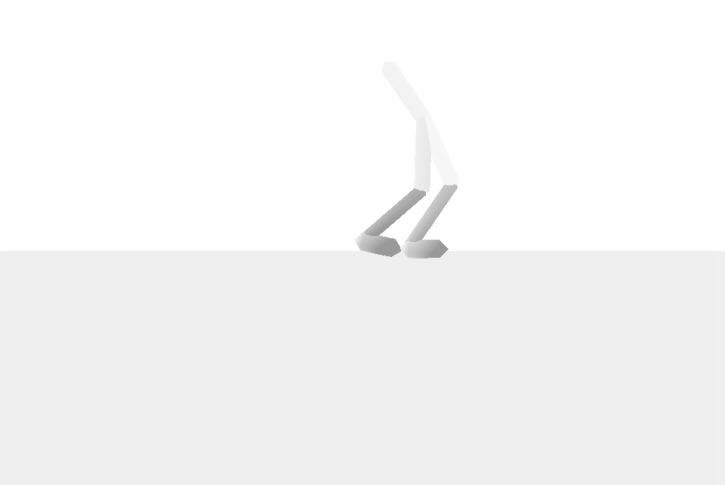}}
    \qquad
    \subfloat[HalfCheetah]{\includegraphics[width=0.2\linewidth]{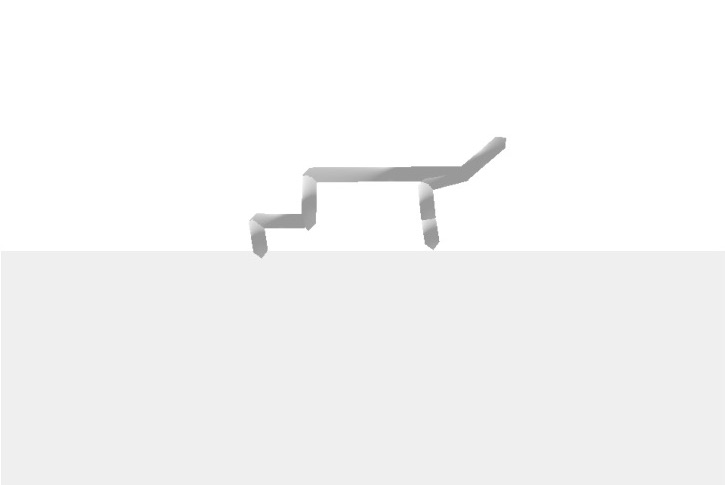}}
    \qquad
    \subfloat[FetchReach]{\includegraphics[width=0.2\linewidth]{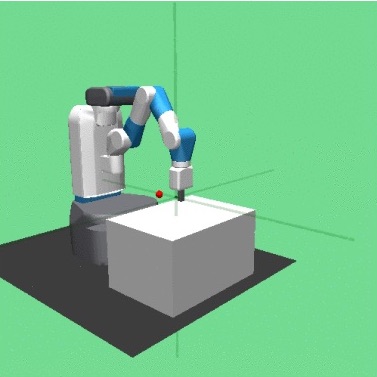}}
    \caption{Simulation environments of RL tasks we are using in this work.}
    \label{fig:rl_environments}
\end{figure*}

\begin{table}[H]
\centering
\caption{Evaluation results after 10 retraining epochs.}
\label{tab:class_retraining_eval}
\begin{tabular}{@{}ccc@{}}
\toprule
Model               & Final test accuracy (\%) & Final test loss (\%) \\ \midrule
New                             & 97.5                    & 9.75                \\
Fused                           & 98.3                    & 7.96                \\
Original 0s                     & 97.9                    & 9.04                \\
Original 1s                     & 97.9                    & 10.1                \\ \bottomrule
\end{tabular}%
\end{table}
Moreover, the final performance of the fused model remains visibly higher than the other models after all 10 epochs have finished. From \autoref{tab:class_retraining_eval}, we can see that the fused model exhibits the highest final test accuracy as well as the lowest final test loss, as evaluated on the testing dataset of 10,000 examples. 

\subsection{Experimental Setup: RL Tasks}
\begin{table*}[t!]
\centering
\caption{Mean episode rewards for each trained parent policy over 10 evaluation episodes.}
\label{tab:parent_policies_eval}
\resizebox{0.75\textwidth}{!}{%
\begin{tabular}{@{}ccccc@{}}
\toprule
Parent Policy & Reacher Task & Walker2d Task & HalfCheetah Task & FetchReach Task \\ \midrule
First & 20.10 $\pm$ 6.80 & 660.80 $\pm$ 420.96 & 2258.43 $\pm$ 642.28 & -2.69 $\pm$ 0.19 \\
Second & 19.57 $\pm$ 10.06 & 728.28 $\pm$ 340.60 & 2013.14 $\pm$ 8.02 & -1.48 $\pm$ 0.07 \\ \bottomrule
\end{tabular}%
}
\end{table*}

To test the effectiveness of the OT technique for policy fusion, we construct four separate RL robotics tasks using OpenAI Gym \cite{openai}: Reacher, Walker2d, HalfCheetah and FetchReach. The first three tasks are simulated using PyBullet \cite{pybullet} while the last task is simulated using Mujoco \cite{mujoco}. Example renders are shown in \autoref{fig:rl_environments}. For each task, the policy of the robotic agent is represented as a Multilayer Perceptron (MLP) with 2 hidden layers of 64 nodes each. The size of the input and output layers differ across the tasks according to \autoref{tab:nn_architecture}, since the environments have varying observation and action spaces. All policies will be trained using the PPO algorithm.

To test each task, we will first independently train two robotic agents on different configurations of the task. Both agents will be trained for 500,000 timesteps. The mean episode rewards, as evaluated on 10 separate and randomly generated environments, are included in \autoref{tab:parent_policies_eval}. These agents will act as the ``parents" of the Renaissance agent, which we will generate via the OT policy fusion method. Then, we will refine the Renaissance agent on a new domain, and compare its performance against retraining the parent agents and training a brand new agent over the same time period. To provide an additional benchmark for OT, we will also compare it against naive averaging as an alternative method of fusing policy weights.

\subsection{Generalising over Changes in the Environment}\label{sec:2d_reacher}

\subsubsection{Experimental Setup}
The learned behaviour of a robotic agent is highly sensitive to changes in the environment, particularly when such changes impact how rewards are earned. To demonstrate that the OT fusion technique is able to generalise the knowledge of two agents trained in differing environment configurations, we construct a simple RL task using the Gym environment Reacher. This task involves a two-link robotic arm as the agent, which operates within a two-dimensional workspace. The goal is for the end-effector to reach to a randomly located target position, with the agent gaining 0.1 rewards for each timestep that it does so. The input layer of the agent's policy network takes in the angular positions and velocities of the robotic arm, and the output layer defines the motor torque of the central joint and the elbow joint. 

The first policy learns to reach to targets randomly located in the first quadrant of the workspace, while the second policy learns to reach to targets randomly located in the second quadrant of the workspace. We now reformulate the problem scope to targets that are placed randomly anywhere within the first and second quadrants. Since the required kinematics of the two-link robotic arm reaching to targets in each quadrant differ significantly, the parent policies have limited capabilities in this new domain. We fuse the parent policies using the OT technique to produce an initialisation of weights for the Renaissance agent. 

\subsubsection{Results}
\begin{figure}[H]
    \centering
    \includegraphics[width=0.8\linewidth]{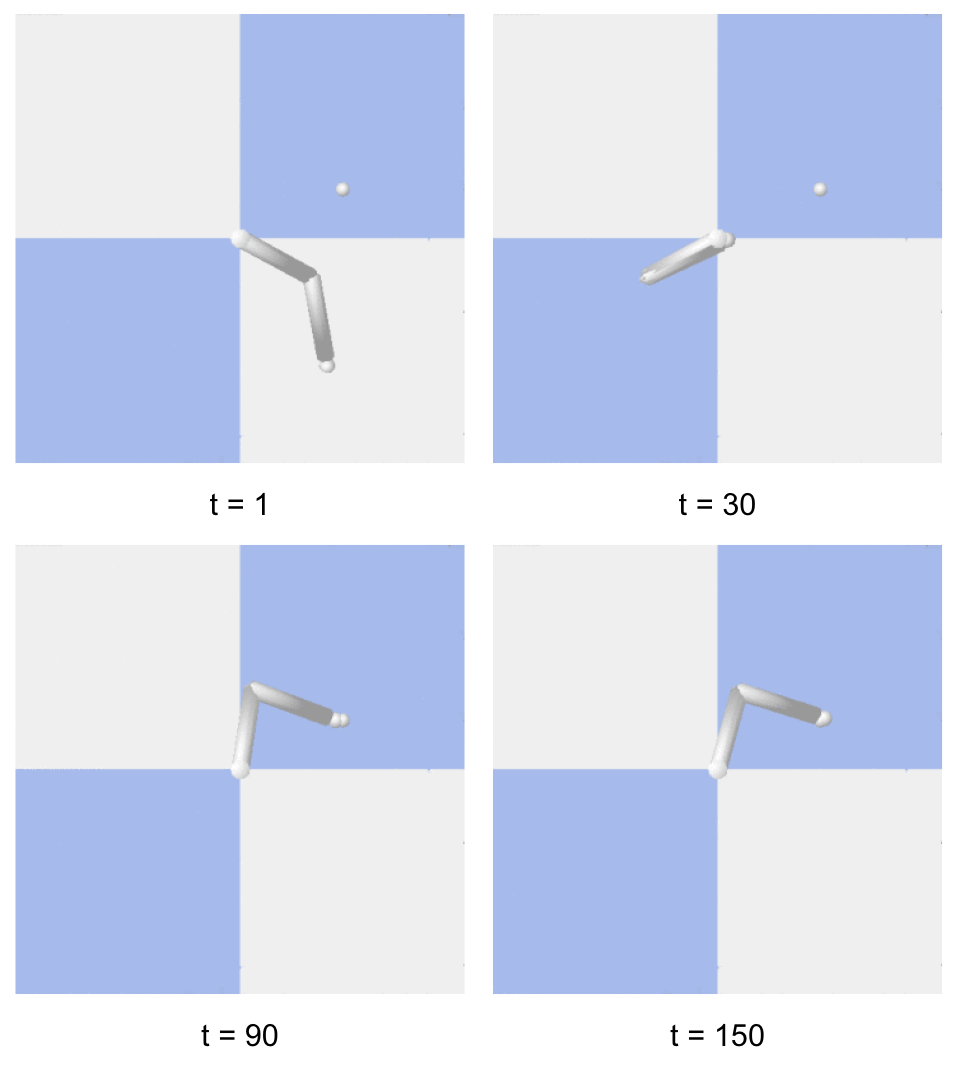}
    \caption{Renders of fused agent after retraining on new Reacher task.}
    \label{fig:new_fused_visualisation}
\end{figure}

After retraining on the new domain for 500,000 timesteps, the fused agent exhibits consistent performance. Renders of the fused agent over a random evaluation episode are shown in \autoref{fig:new_fused_visualisation}.

\begin{figure}[H]
    \centering
    \includegraphics[width=\linewidth]{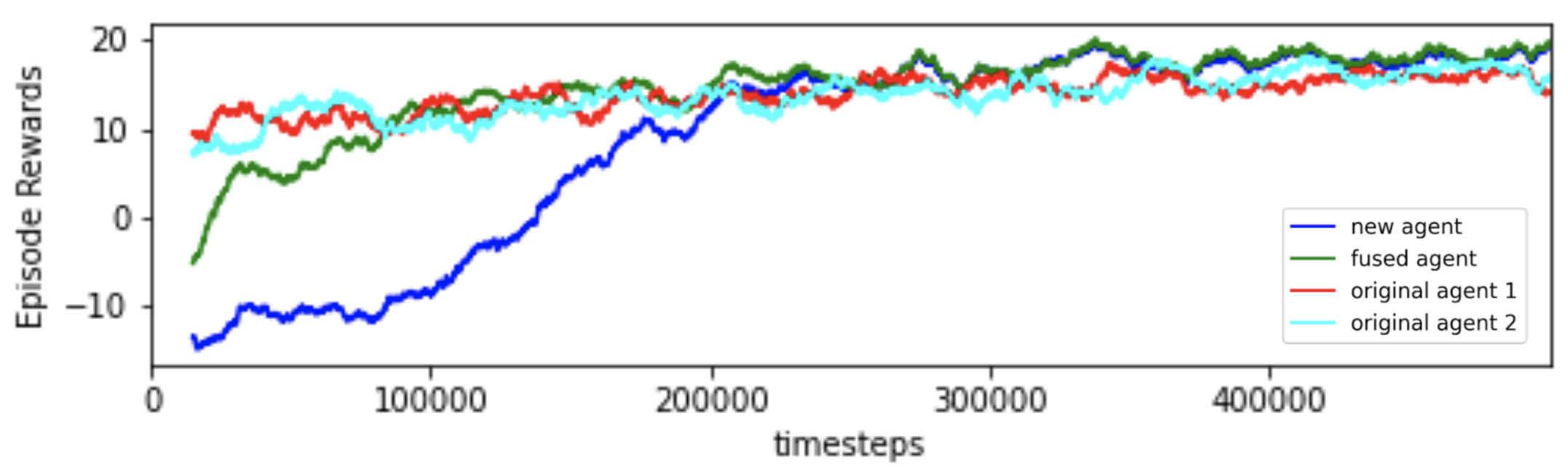}
    \caption{Retraining curves for Reacher environment.}
    \label{fig:rl_retraining}
\end{figure}

The mean episode rewards for the fused agent, compared with the other benchmark agents, are shown in \autoref{fig:rl_retraining}. It is evident that the fused agent achieves high consistent performance much faster than the rest, surpassing 15 rewards per episode after 200,000 timesteps. In contrast, it takes approximately 350,000 timesteps for the next best agent to converge to similar performance. Upon completion of the retraining period, we can see from \autoref{tab:rl_retraining_eval} that the fused agent exhibits the highest mean reward as well as the lowest standard deviation, as tested over 10 random evaluation episodes. 

\begin{table*}[t!]
\centering
\caption{Mean episode rewards over 10 evaluation episodes.}
\label{tab:rl_retraining_eval}
\resizebox{0.75\textwidth}{!}{%
\begin{tabular}{@{}ccccc@{}}
\toprule
Policy & Reacher Task & Walker2d Task & HalfCheetah Task & FetchReach Task \\ \midrule
New Agent & 16.64 $\pm$ 9.51 & 972.02 $\pm$ 5.88 & 2244.47 $\pm$ 898.70 & -5.38 $\pm$ 4.03 \\
\textbf{Fused Agent} & \textbf{17.05 $\pm$ 7.37} & \textbf{1520.59 $\pm$ 232.58} & \textbf{2810.22 $\pm$ 29.93} & \textbf{-5.05 $\pm$ 3.4} \\
Original Agent 1 & 15.17 $\pm$ 8.03 & 1000.34 $\pm$ 24.44 & 2750.57 $\pm$ 52.82 & -7.13 $\pm$ 3.73 \\
Original Agent 2 & 16.59 $\pm$ 7.92 & 822.42 $\pm$ 345.61 & 2500.23 $\pm$ 444.02 & -5.95 $\pm$ 3.69 \\
Naive Averaging & 15.09 $\pm$ 8.08 & 580.29 $\pm$ 420.92 & 1853.18 $\pm$ 1312.92 & -8.03 $\pm$ 3.47 \\ \bottomrule
\end{tabular}%
}
\end{table*}

\subsection{Generalising over Changes in Robot Kinematics}

\subsubsection{Experimental Setup}
Agents with different robot kinematics will learn different behaviours to reach the desired goal, even if the environment is the same. To demonstrate that the Renaissance agent generalises over different robot kinematics, we construct two RL tasks using the Gym environments Walker2d and HalfCheetah.

\subsubsection{Walker2d}
The Walker2d task involves a bipedal robot as the agent, which operates within a 2D plane (i.e. can only move forwards or backwards). The goal is for the agent to walk forward as fast as possible, with the agent gaining rewards as a function of the current forward velocity, plus a constant bonus for not falling over. The input layer of the agent's policy network takes in the angular positions and velocities of the joints, and the output layer defines the signals to move the torso-connected foot and the left foot.

The first agent is trained with a more powerful torso-connected foot (coefficient = 40) compared to its left foot (coefficient = 10), while the second agent is trained with a more powerful left foot (coefficient = 25) compared to its torso-connected foot (coefficient = 20). We now fuse the parent policies and reformulate the target domain to learning the Walker2d task on an agent with equally powerful torso-connected and left feet (coefficient = 30).

\begin{figure}[H]
    \centering
    \includegraphics[width=\linewidth]{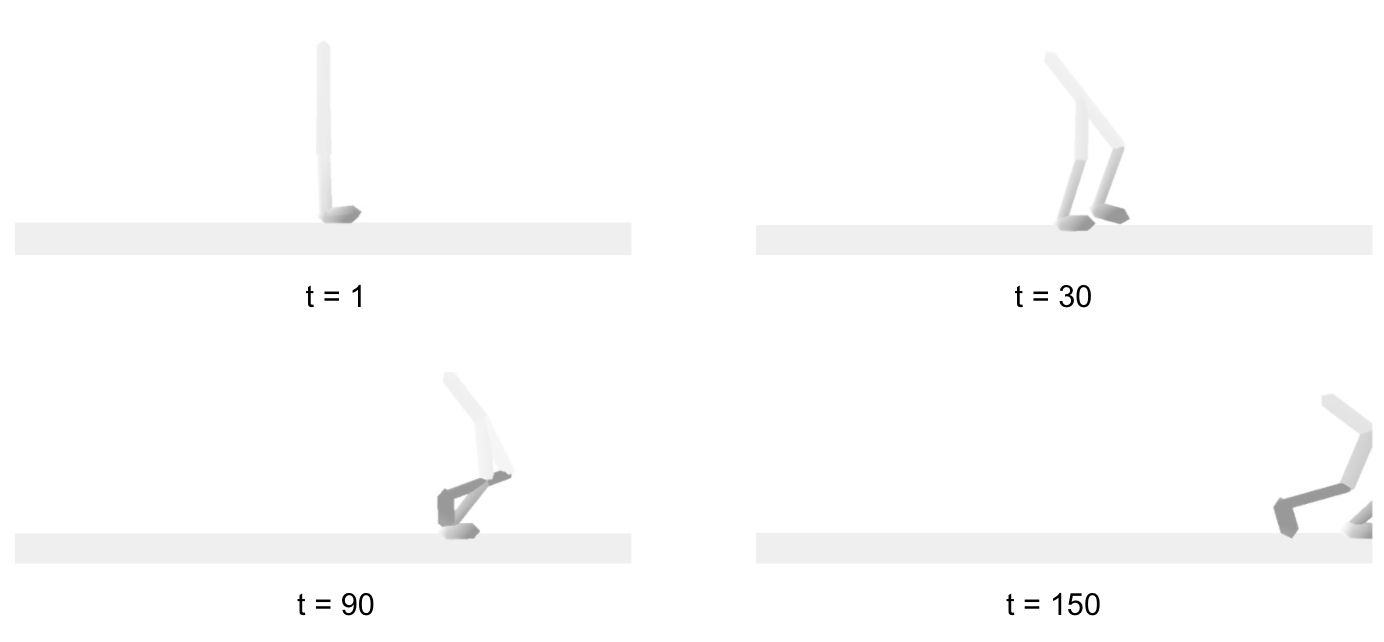}
    \caption{Renders of fused agent after retraining on new Walker2d task.}
    \label{fig:retrained_walker}
\end{figure}

After retraining on the new domain for 500,000 timesteps, the fused agent is able to walk forward in a natural-looking manner. Renders of the agent over a random evaluation episode are shown in \autoref{fig:retrained_walker}.

\begin{figure}[H]
    \centering
    \includegraphics[width=\linewidth]{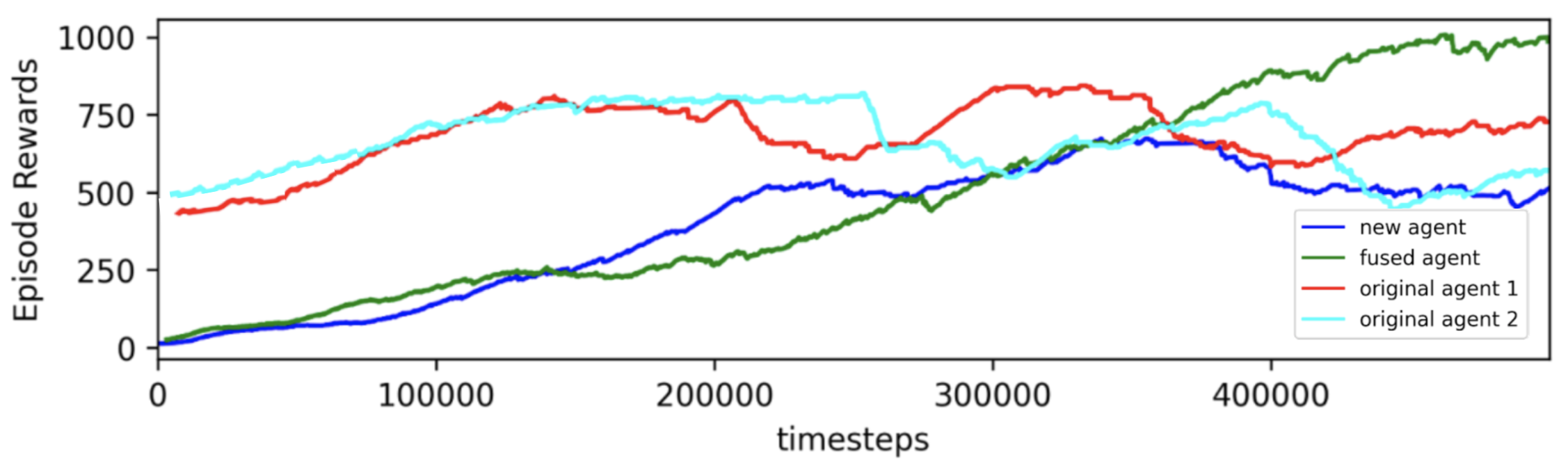}
    \caption{Retraining curves for Walker2d environment.}
    \label{fig:rl_retraining_walker}
\end{figure}

The mean episode rewards for the fused agent, compared with the other benchmark agents, are shown in \autoref{fig:rl_retraining_walker}. We can see that the fused agent is the only one that achieves high consistent performance, surpassing 1000 rewards per episode by the end of the retraining period. From \autoref{tab:rl_retraining_eval}, it is evident that the fused agent exhibits the highest minimum, maximum and mean reward, as tested over 10 random evaluation episodes.

\subsubsection{HalfCheetah}
We further demonstrate the ability of OT policy fusion to generalise over different robot kinematics by extending to a more complex agent. The HalfCheetah task involves a quadruped robot as the agent, which also moves only forwards or backwards within a 2D plane. Similar to the Walker2d task, the goal is for the agent to run forward as fast as possible. The agent's policy network determines how much torque to apply on its thighs, shins and feet.

We train the first parent policy on an agent with greater thigh and shin power, but reduced foot power. This results in more rigid foot joints. We then train the second parent policy on an agent with the opposite kinematics, resulting in more rigid thigh and shin joints. Similar to the Walker2d task, we now fuse the policies and change the target domain to learning the HalfCheetah task on an agent with neutral power for all joints. This should allow the agent to learn joint movements that are less rigid.

\begin{figure}[H]
    \centering
    \includegraphics[width=\linewidth]{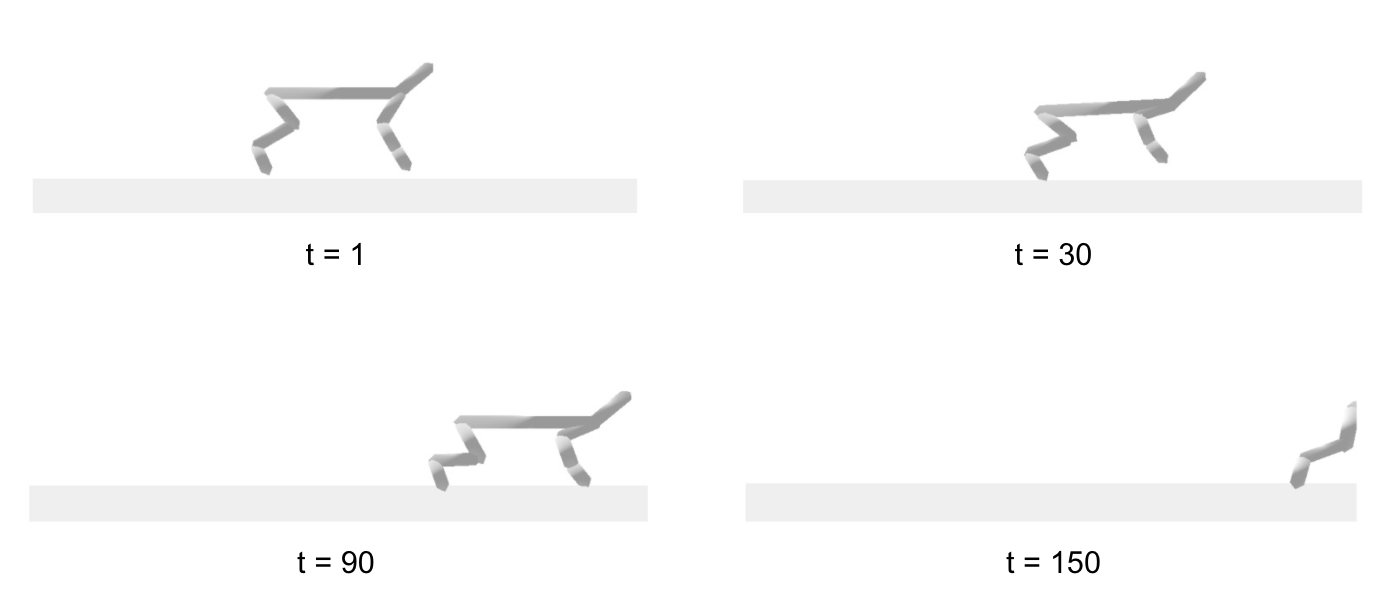}
    \caption{Renders of fused agent after retraining on new Cheetah task.}
    \label{fig:retrained_cheetah}
\end{figure}

After retraining on the new domain for 500,000 timesteps, the fused agent is able to run forward without any of the joint rigidity displayed by the parent agents. Renders of the agent over a random evaluation episode are shown in \autoref{fig:retrained_cheetah}.

\begin{figure}[H]
    \centering
    \includegraphics[width=\linewidth]{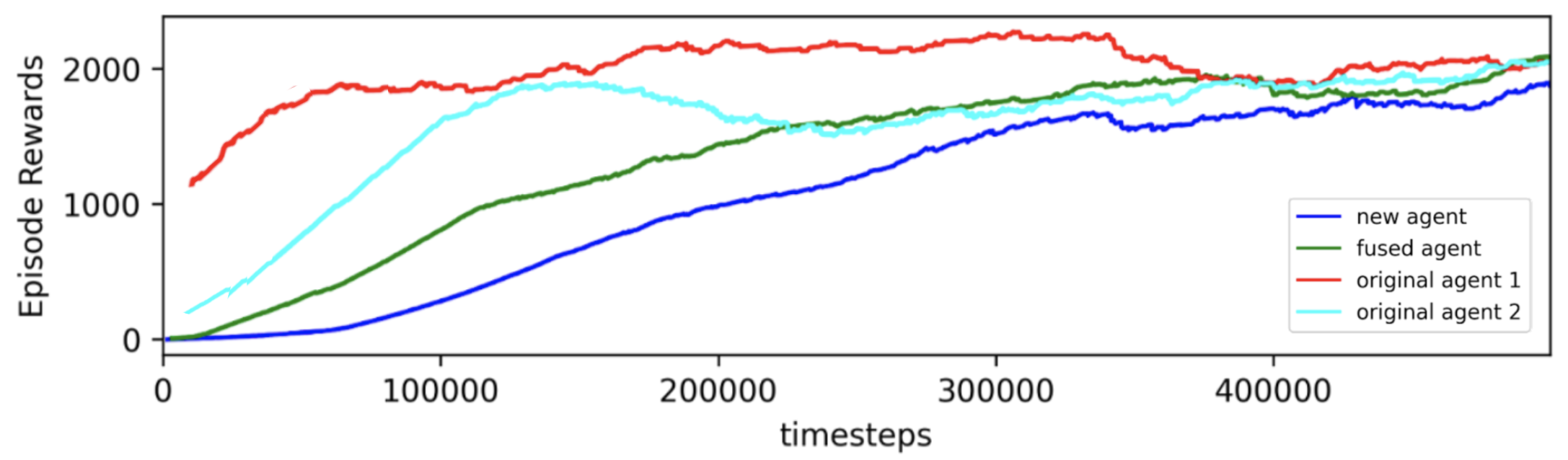}
    \caption{Retraining curves for HalfCheetah environment.}
    \label{fig:rl_retraining_cheetah}
\end{figure}

The mean episode rewards for the fused agent, compared with the other benchmark agents, are shown in \autoref{fig:rl_retraining_cheetah}. It is evident that the fused agent learns much faster than the new agent, and overtakes the parent agents by the end of the retraining period. From \autoref{tab:rl_retraining_eval}, we can further observe that the fused agent exhibits the highest mean reward and lowest standard deviation, as tested over 10 random evaluation episodes.

Additionally, the retrained parent agents still exhibit very similar kinematics to what they learned from their original training domain. In contrast, both the fused agent and the new agent display the desired kinematics of the new domain, which has less joint rigidity. This demonstrates how the OT fusion technique is useful for consolidating the common knowledge of parent models which have been over-optimised to a previous task.

\subsection{Evaluating on Robotic Arm Task}

\subsubsection{Experimental Setup}
The ``Fetch" Gym environments are based on the 7-DoF Fetch Robotics arm. This provides a more complex agent, observation and action space compared to the Reacher task shown in \autoref{sec:2d_reacher}. We extend that task by using the Gym environment FetchReach.

The FetchReach task involves the Fetch arm as the agent, which operates within a 3D space. The goal is for the agent to move its end-effector to target positions randomly located above a workspace. Rewards are binary: at each timestep, the agent obtains a reward of 0 if its end-effector is at the target position, and -1 otherwise. The agent's policy network takes in the Cartesian positions and linear velocities of the arm links and gripper as observations, and outputs the desired movements in Cartesian coordinates as actions.

In the previous 2-link Reacher task, we set the target locations to be within the first or second quadrants of the workspace. However, since the Fetch robot stands above the workspace and moves in 3D, these existing problem formulations are not different enough to ensure that the parent policies will learn distinct behaviours. Instead, the first policy learns to move the end-effector to target positions located over the bottom left of the workspace, while the second policy learns for the top right of the workspace. We now fuse the policies and change the problem scope to target positions that are located randomly anywhere over the workspace, excluding the corners.

\begin{figure}[H]
    \centering
    \includegraphics[width=0.9\linewidth]{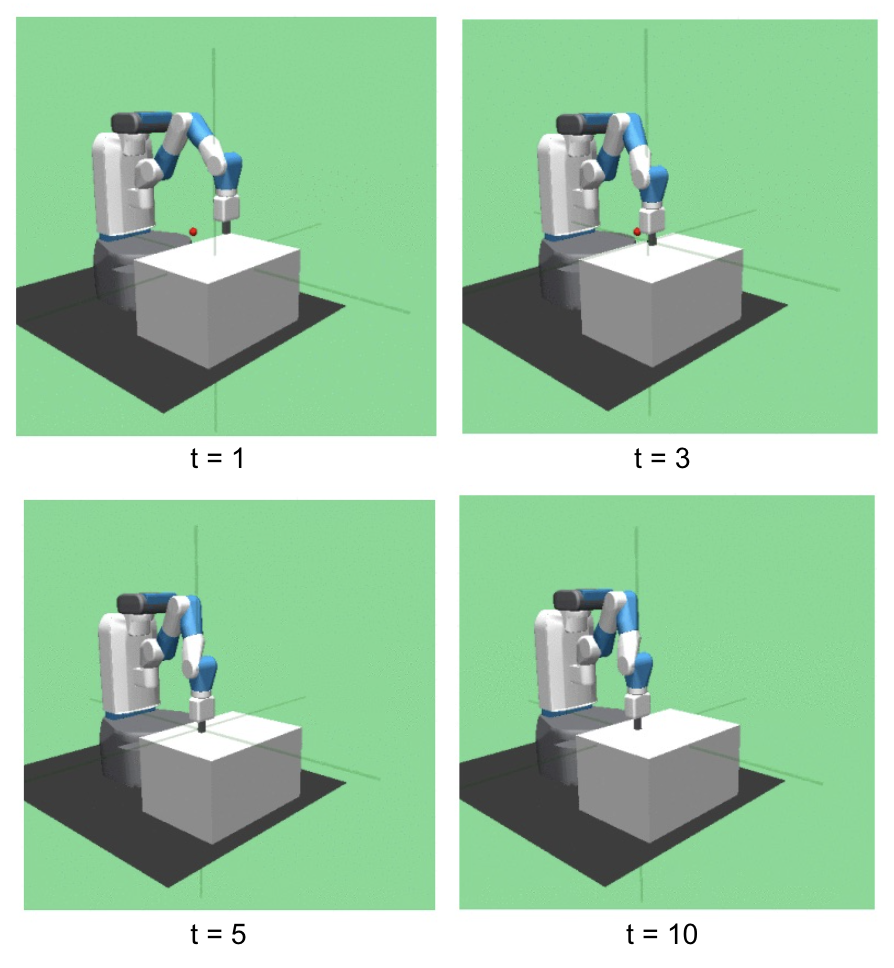}
    \caption{Renders of fused agent after retraining on new FetchReach task.}
    \label{fig:retrained_fetch_reach}
\end{figure}

\subsubsection{Results}
After retraining on the new domain for 200,000 timesteps, the fused agent is able to quickly and consistently move its end-effector to target positions located anywhere over the central area of the workspace. Renders of the agent over a random evaluation episode are shown in \autoref{fig:retrained_fetch_reach}.

\begin{figure}[H]
    \centering
    \includegraphics[width=\linewidth]{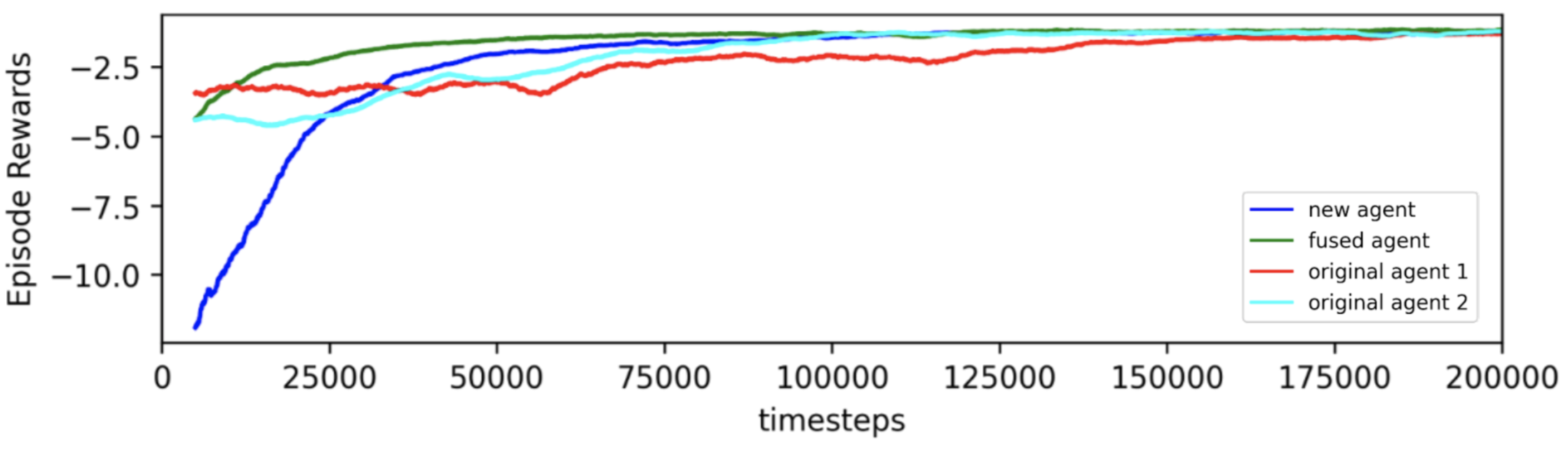}
    \caption{Retraining curves for FetchReach environment.}
    \label{fig:rl_retraining_fetch_reach}
\end{figure}

The mean episode rewards for the fused agent, compared with the other benchmark agents, are shown in \autoref{fig:rl_retraining_fetch_reach}. It is evident that the fused agent achieves high consistent performance much faster than the rest, surpassing -2.5 rewards per episode by 25,000 timesteps. In contrast, it takes approximately 100,000 timesteps for the next best agent to converge to similar performance. From \autoref{tab:rl_retraining_eval}, we can see that the fused agent achieves the highest mean reward as well as the lowest standard deviation, as tested over 10 random evaluation episodes.

%% file: conclusion.tex
In this work, we developed policy fusion utilising OT theory to create Renaissance agents. We demonstrated that with minimal retraining, the Renaissance agents can quickly learn a new domain by leveraging generalised prior knowledge. In contrast, the parent agents are often over-optimised to their original tasks, and thus less adept at learning diverse skills in a variety of common RL robotics problems. Ultimately, the fused agents were empirically demonstrated to learn new domains significantly faster and achieve superior sustained performance compared with other benchmarks. This is especially valuable for RL in robotics since training is highly time-consuming and resource expensive. In light of the simulation-to-simulation examples discussed in this work, applying the proposed technique to simulation-to-real policy fusion is a promising future direction.